\pdfoutput=1

\documentclass[11pt]{article}
\usepackage{amsfonts}
\usepackage{dashrule}
\usepackage[final]{acl}
\usepackage{appendix}
\usepackage{tabularx}
\usepackage{xcolor}
\usepackage{amsmath}
\usepackage{booktabs}
\usepackage{multirow} 
\usepackage{colortbl}
\usepackage{times}
\usepackage{arydshln}
\usepackage{fancyhdr}
\usepackage{xcolor}
\usepackage{placeins}
\usepackage{afterpage}
\usepackage{float}
\usepackage{latexsym}
\usepackage{graphicx}
\usepackage[T1]{fontenc}

\usepackage[utf8]{inputenc}
\usepackage{microtype}

\title{Citation-Enhanced Generation for LLM-based Chatbots }

\author{
\textbf{Weitao Li$^{1,2}$}, \textbf{Junkai Li$^{1,2}$}, \textbf{Weizhi Ma$^{2,\dagger}$}, \textbf{Yang Liu$^{1,2,3,\dagger}$}\\
$^1$ Dept. of Comp. Sci. \& Tech., Institute for AI, Tsinghua University, Beijing, China \\
$^2$ Institute for AI Industry Research (AIR), Tsinghua University, Beijing, China \\
$^3$ Jiangsu Collaborative Innovation Center for Language Competence, Jiangsu, China\\
}
\begin{document}
\thispagestyle{fancy}
\fancyhf{}
\rhead{Accepted at ACL 2024 Main Conference}
\maketitle

\begin{abstract}
Large language models (LLMs) exhibit powerful general intelligence across diverse scenarios, including their integration into chatbots. However, a vital challenge of LLM-based chatbots is that they may produce hallucinated content in responses, which significantly limits their applicability. Various efforts have been made to alleviate hallucination, such as retrieval augmented generation and reinforcement learning with human feedback, but most of them require additional training and data annotation. In this paper, we propose a novel post-hoc \textbf{C}itation-\textbf{E}nhanced \textbf{G}eneration (\textbf{CEG}) approach combined with retrieval argumentation. Unlike previous studies that focus on preventing hallucinations during generation, our method addresses this issue in a post-hoc way. It incorporates a retrieval module to search for supporting documents relevant to the generated content, and employs a natural language inference-based citation generation module. Once the statements in the generated content lack of reference, our model can regenerate responses until all statements are supported by citations. Note that our method is a training-free plug-and-play plugin that is capable of various LLMs. Experiments on various hallucination-related datasets show our framework outperforms state-of-the-art methods in both hallucination detection and response regeneration on three benchmarks. Our code and datasets can be found at \url{https://github.com/Tsinghua-dhy/CEG}. 
\end{abstract}



\maketitle
{\let\thefootnote\relax\footnotetext{$\dagger$ Weizhi Ma (mawz@tsinghua.edu.cn) and Yang Liu (liuyang2011@tsinghua.edu.cn) are corresponding authors.}}

\section{Introduction}
Large Language Models (LLMs) have experienced rapid development in recent years, which show powerful general intelligence in various scenarios~\cite{yue2023disc,singhal2023towards}.  Current LLM-based chatbots, epitomized by ChatGPT and GPT-4, demonstrate impressive capabilities across distinct domains in communicating with humans. 
There is a growing consensus that LLM-based chatbots can be the next generation of information acquisition methodology.

\begin{figure}[t!]
    \centering
    \includegraphics[width=\columnwidth]{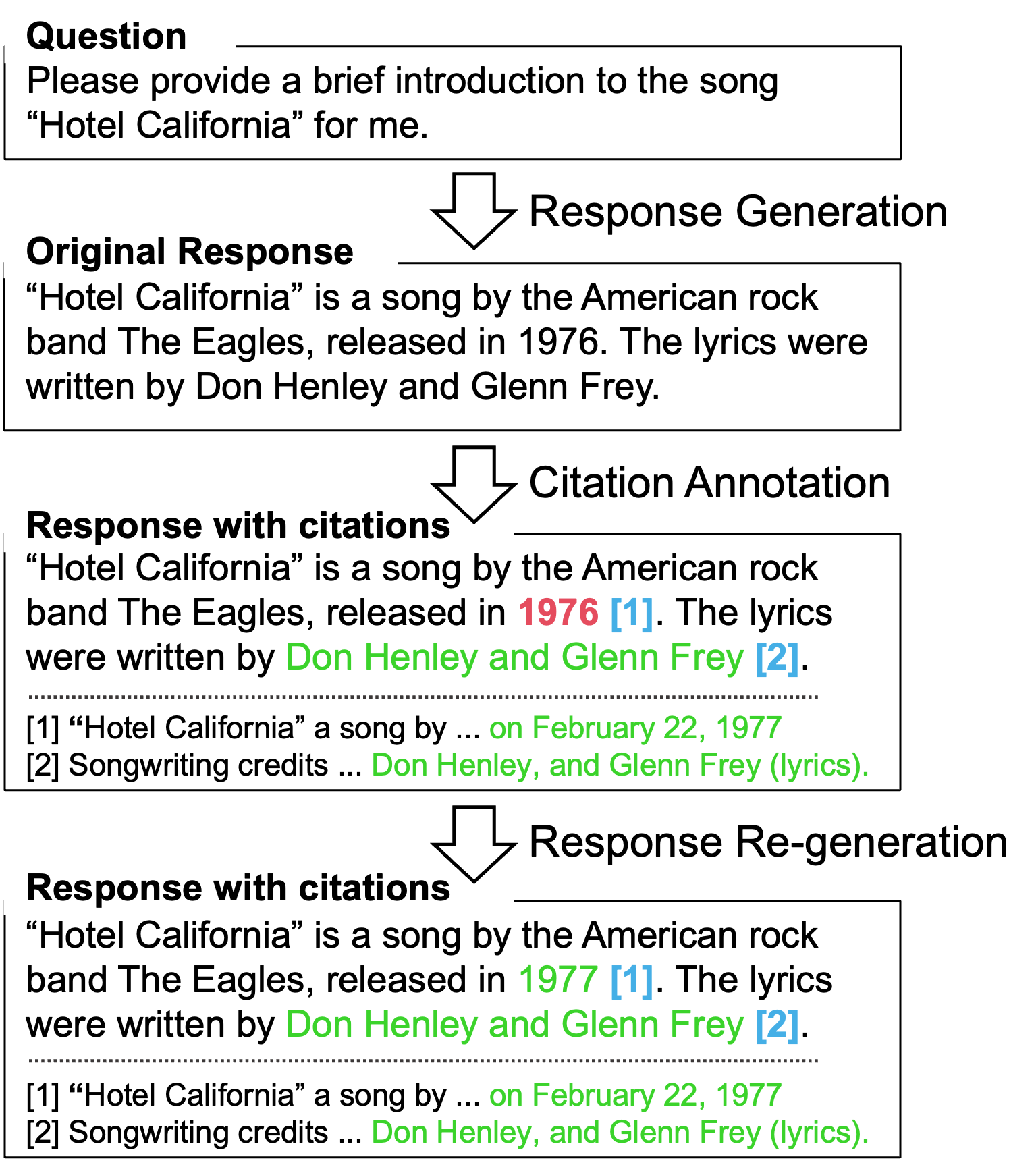}
    \caption{An illustration of our method, which adds citations for the generated content. If there are hallucinations in the generated content, we prompt LLM to regenerate a new response. }

    \label{fig:intro}
\end{figure}

However, a critical and unsolved challenge of LLM-based chatbots is the hallucination problem~\cite{ji2023survey}, which indicates these chatbots may generate hallucinated content in responses randomly. As the underlying mechanisms of hallucinations remain unclear, this problem has substantial constraints on the deployment of LLM-based chatbots in various sensitive scenarios, such as healthcare and education, where reliability is paramount.

Previous approaches have attempted to mitigate this issue through retrieval augmentation~\cite{borgeaud2022improving,izacard2022few} and value alignment (RLHF)~\cite{ouyang2022training,touvron2023llama} in response generation, but these often require additional training and extensive data annotation. For example, InstructGPT~\cite{ouyang2022training} utilize RLHF to alleviate hallucinations in model output, but needs extra training. \citet{gao-etal-2023-enabling} attempt to reduce hallucination through adding retrieved related documents and citations before generation, while the pre-hoc way of incorporating citations may potentially harm the model performance, resulting in poor response results with hallucinations.

In this work, we propose a novel method to alleviate hallucination in LLMs, which leverages retrieval augmentation and Natural Language Inference (NLI) technologies to implement Citation-Enhanced Generation (CEG) in a post-hoc way. Figure~\ref{fig:intro} is an illustration.
Differing from previous studies, the retrieval augmentation module of the CEG framework works after generation (post-hoc), and CEG prompts the model to regenerate the answer when necessary. This approach is effective and easy to use, which can reduce the hallucination in the model's output for various LLMs. 
We conduct experiments on distinct hallucination-related benchmarks, including detection and response regeneration, where our method achieved state-of-the-art performance. Further analyses demonstrate the usefulness of each module on CEG. 

In summary, the main contributions of our work can be summarized as follows:
\begin{itemize}
    \item We are the first to propose the use of citation to alleviate hallucination in a post-hoc way with regeneration.
    \item We design a novel post-hoc citation-enhanced generation framework combined with retrieval augmentation and NLI to avoid hallucinations, which is flexible for existing LLMs.
    \item Experimental results show that our CEG framework achieves the best performance on three hallucination-related benchmarks.
\end{itemize}

\section{Related Work}
\subsection{Hallucination Control in LLMs}
Generative AI has achieved significant advancements, while still facing the hallucination problem. Existing strategies can be categorized into major two types: mitigation during training and mitigation during inference. For the first type, LLMs, such as LLaMA 2~\cite{touvron2023llama}, undergo extensive training cycles with high-fidelity data sources like Wikipedia to bolster factual consistency in pre-training. \citet{zhou2023lima} alleviate hallucination during instruction fine-tuning, which adopts high quality manually annotated content to regulate hallucination. Some studies~\cite{ouyang2022training,touvron2023llama} also introduce penalties for nonfactual responses to alleviate hallucination in RLHF. However, all these methods need extra training and annotations. 

On the other hand, researchers try to deal with the hallucination challenge during inference. 
Inference-Time-Intervention~\cite{li2023inference} mitigates hallucination by shifting model activations along these factuality-related directions during inference. Retrieval-Augmented Generation~(RAG)~\cite{lewis2020retrieval} has become a prevalent technique in alleviating hallucination by retrieving reliable documents before generation~\cite{yu-2022-retrieval}. While these methods still generate hallucinations due to the lack of post-hoc verification and they are unable to provide citations for verification. 

\subsection{Citation Augmented LLMs}
In the realm of LLMs, retrieval technology has become a crucial component~\cite{zhang2023siren,gao2023retrieval}, as it provides related knowledge in generating more reliable results (also mitigates the occurrence of hallucinations). Previous studies point out that citation, generated by retrieval models, is key to building responsible and accountable LLMs~\cite{huang2023citation}. 

Existing citation augmented strategies can be divided into two types: parametric and non-parametric. Parametric methods~\cite{taylor2022galactica} refer to information internalized from the training data, often leading to inaccurate annotated documents, as the annotation process itself can give rise to hallucinations. 
Non-parametric methods~\cite{gao-etal-2023-enabling,menick2022teaching,izacard-grave-2021-leveraging} involve querying relevant information and seamlessly integrating the retrieved content from outside corpus, which provides more reliable citations. Thus, most previous studies are non-parametric, but they are pre-hoc based. For example, \citet{gao-etal-2023-enabling} adopt retrieval processes to facilitate the annotation of documents within model-generated outputs. Nevertheless, their pre-hoc annotation strategy inadvertently escalates the complexity of a QA task by converting it into a dual challenge of generating a response coupled with simultaneous annotation. 
Different from existing citation augmented studies, we propose a different strategy to utilize retrieval models to generate citations in a post-hoc way.


\begin{figure*}[t!]
\centering
\includegraphics[width=\linewidth]{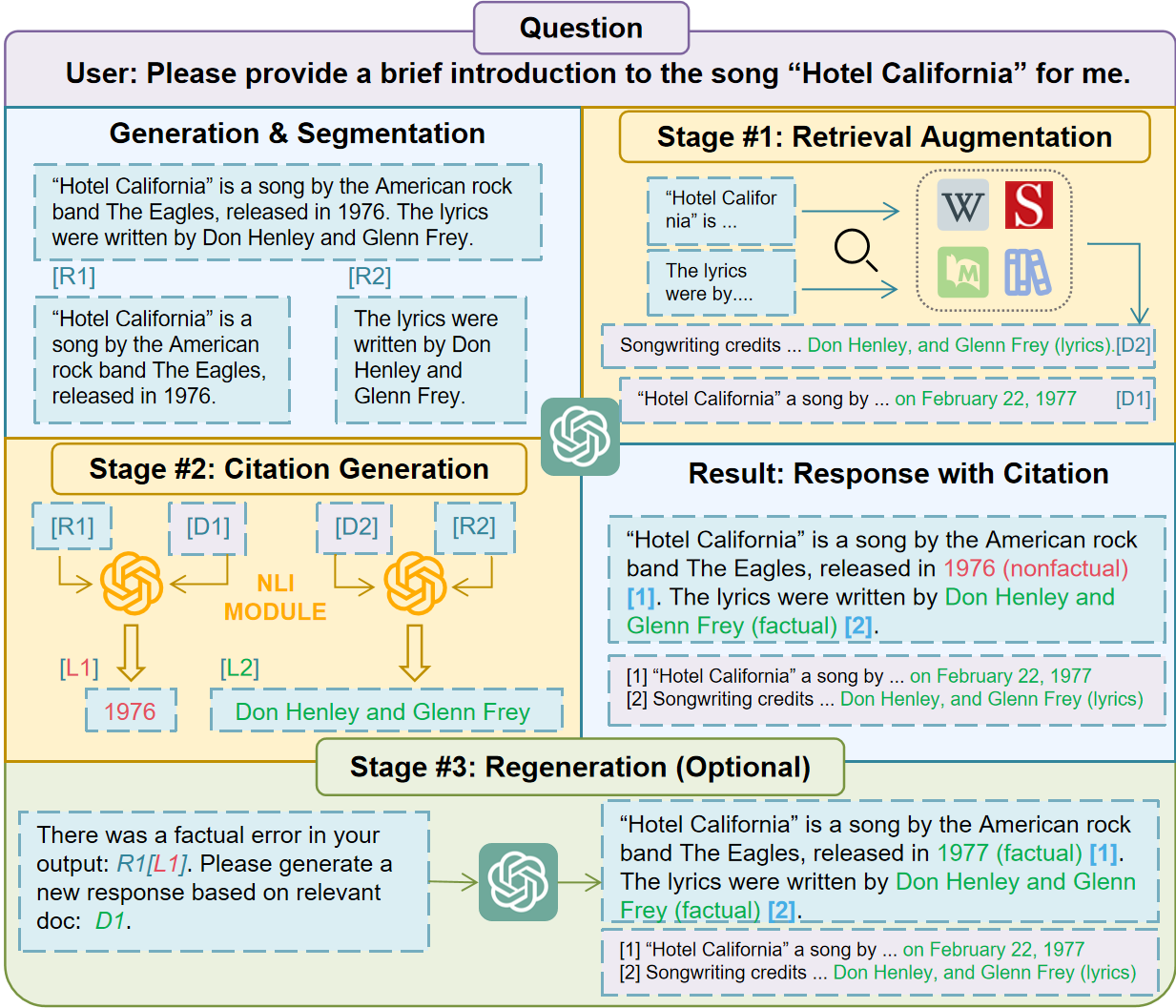}
\caption{An overview of our CEG framework. [R1] and [R2] denote  segments. [D1] and [D2] represent retrieved documents for each segment. [L1] and [L2] are  labels (Factual/Nonfactual) generated by the NLI module.}
\label{fig:2}
\end{figure*}




\section{Approach}
\subsection{Overview}
Firstly, we give an overview of our proposed CEG framework. Illustrated in Figure~\ref{fig:2}, CEG has several critical modules: 1) Retrieval augmentation module, designed to search for documents $D_j$ relevant to the original response $R$. In cases where responses are excessively lengthy, they can be broken down into sub-claims $R = \{R_1, R_2, ..., R_n\}$. 2) Citation generation module, which assesses if the retrieved documents $D_j$ substantiate the \{$R_i$\} in response or not. 3) Regeneration module, tasked with creating a new prompt that integrates the original user query and key retrieved information for the LLM $M$ to get a more reliable response $R'$. 

It is important to note that our method is a post-hoc framework and is highly adaptable across different LLMs, as it does not require any additional training or fine-tuning. Consequently, we do not specify a particular LLM here.


\subsection{Retrieval Augmentation Module}

Retrieval augmentation has been shown to have powerful abilities in previous hallucination-related studies~\cite{gao-etal-2023-enabling,zhao-etal-2023-verify}. Different from these studies that aim to retrieve documents as evidence before response generation (questions are queries), we propose to conduct retrieval augmentation in a post-hoc way to verify the correctness of the generated claim $R_i$ (claims are queries). 
As there are various existing studies on how to retrieve the most related document, we use a simple but effective dense retrieval strategy to verify the performance of our CEG framework, and we believe stronger retrieval will bring further improvements.

\textbf{Query}: For the response $R$, it will be segmented into several claims if necessary, resulting in $R = \{R_1, R_2, ..., R_n\}$.  
Here, we adhere to previous work~\cite{chen2023felm} and employ a heuristic algorithm for segmentation using the NLTK~\cite{bird2009natural} sentence tokenizer. The NLTK sentence tokenizer is a well-performing and widely used ~\cite{chen2023felm,bird2009natural,liu2023argugpt} sentence tokenizer and generally segments text correctly in most cases. We split $R$ to obtain reasonable results that align with user reading habits to get the claims $R_i$.  Then, $R_i$ is adopted as the query one by one.

\textbf{Corpus (Candidate Documents)}: The choice of Corpus decides the scope of applications, and there are multiple candidates. In this study, our focus lies predominantly in the domain of knowledge-based question answering, necessitating the employment of a curated corpus. To this end, we leverage a processed snapshot of Wikipedia from October 20, 2023\footnote{https://dumps.wikimedia.org/enwiki/}, segmented into approximately 100-word  candidate documents, each demarcated by a period or newline character. Note that you can replace it with any other corpus, and we use it as most hallucination benchmarks are based on Wikipedia.

\textbf{Retriever}: Dense vector based retrieval technologies have demonstrated powerful performances in recent years, which are also widely used in existing RAG models. Here, we adopt the SimCSE BERT\footnote{https://huggingface.co/princeton-nlp/sup-simcse-bert-base-uncased/tree/main}~\cite{gao-etal-2021-simcse,wang-etal-2023-collective} as the query and document encoder for its promising efficiency in previous studies~\cite{wang-etal-2023-self-knowledge,wang-etal-2023-collective}. Then, candidate documents are ranked based on cosine similarity scores calculated by the following equation: 
 \[\text{Sim}(R_i, d_j) = \frac{e(R_i) \cdot e(d_j)}{\|e(R_i)\| \cdot \|e(d_j)\|},\]
Where $e(·)$ is the SimCSE BERT encoder, $d_j$ is a candidate document in the corpus. 
As more documents need more calculation in further modules, the top-$k$ retrieved documents with higher similarity are selected to form the  reference document set \(D_i\). We add an extra threshold $t$ to filter out the retrieved documents that have low cosine similarity. Apart from the top-$1$ document, if the Sim(\(R_i, d_j\)) < $t$, \(d_j\) will not be included in \(D_i\). These documents are subsequently concatenated to construct the final retrieved content $D_i$ for further calculation. 

\subsection{Citation Generation Module}

After getting the reference document $D_i$ for each response segment $R_i$, the next step involves generating labels and citations to verify the correctness of $R_i$. We propose to adopt an NLI method to determine the relationship between each claim-document pair ($R_i$, $D_i$). In general, the relationship can be categorized into three types: support, independence, and contradiction. But in hallucination-related scenarios, to adhere to previous studies, we only utilize two categories: (1) \textbf{Factual}, where $D_i$ serves as a reference for $R_i$, thereby substantiating the claim. (2) \textbf{Nonfactual}, which means $D_i$ presents a opposite claim to $R_i$.

Although numerous models~\cite{honovich-etal-2022-true-evaluating,raffel2020exploring} are capable of the NLI method, our CEG framework seeks to fully leverage the language comprehension capabilities of LLMs. Therefore, we prefer to utilize an LLM with predefined prompts to serve as the NLI method. An illustrative prompt is provided below:

\noindent\makebox[\linewidth]{\hdashrule{\linewidth}{1pt}{1mm}}
\textbf{Instruction}: I will show you a \textcolor[HTML]{0000ff}{question}, a \textcolor[HTML]{ff7f00}{response segment} of this question, and a \textcolor[HTML]{0abf53}{reference document}. Your task is to assess whether the given \textcolor[HTML]{ff7f00}{response segment} contains \textcolor[HTML]{e54c5e}{factual errors} or not with the help of the \textcolor[HTML]{0abf53}{reference document}. ... 
\noindent\makebox[\linewidth]{\hdashrule{\linewidth}{1pt}{1mm}}

When the LLM output indicates ``factual'', the document $D_i$ is identified as a valid reference for the claim $R_i$. Consequently, this citation can be added to the original response. If none of the retrieved top-k documents substantiate the claim $R_i$ or if there are documents opposing the claim, we will label this claim as nonfactual (potential hallucination) to remind users to keep carefully reading. Based on the introduced two modules, we can detect whether there are hallucinations in responses.


\subsection{Response Regeneration Module}
\label{sec:3.4}
In previous modules, our framework offers a post-hoc method to conduct citation-enhanced verification for responses, where reliable responses are incorporated with citations. However, a new challenge is how to deal with  potential hallucinations. So we propose a response regeneration module.

Assuming LLM $M$ generates the original response $R$, our framework will provide a new prompt for regeneration. The prompt not only contains the original query, but is also incorporated with retrieved documents and the annotated nonfactual segments. Here we provide a brief illustration of the prompt (a full prompt is shown in appendix):

\noindent\makebox[\linewidth]{\hdashrule{\linewidth}{1pt}{1mm}}

\noindent \textbf{User}: \textcolor[HTML]{0000ff}{Question}; \textbf{Chatbot}: \textcolor[HTML]{ff7f00}{Response}; \textbf{User}: [There were some factual errors in your output: (\textit{\textcolor[HTML]{e54c5e}{Nonfactual Claims}}). Please generate a new response based on relevant docs: (\textit{\textcolor[HTML]{0abf53}{Relevant Docs}}).]  

\noindent\makebox[\linewidth]{\hdashrule{\linewidth}{1pt}{1mm}}

Upon receipt of the regenerated response, we can initiate a new citation-enhanced generation process. If the response is adjudged to be free of factual errors, it becomes the final response and will be shown to users. However, if the new response still contains hallucinations, the regeneration cycle will be repeated. 
To conserve API resources and reduce the waiting time, a predefined parameter $T$ can be set to constrain the max regeneration attempts.

\section{Experimental Settings}
\subsection{Overview}
To verify the effectiveness of our framework, we adopt four hallucination-related datasets in our experiments:  WikiBio GPT-3~\cite{manakul-etal-2023-selfcheckgpt}, FELM~\cite{chen2023felm}, HaluEval~\cite{li-etal-2023-halueval}, and WikiRetr. WikiBio GPT-3 and FELM are hallucination detection benchmarks. HaluEval is a hallucination generation benchmark.
Besides, we construct a new dataset named WikiRetr, which is to evaluate the retrieval and citation annotation performance. Due to the tasks and baselines are distinct in various datasets, we will introduce each dataset and corresponding settings one by one.

We use GPT models as the LLM backbones, and the version involved in different datasets is distinct for fair comparison with existing baselines. Unless otherwise specified, ``ChatGPT'' refers to GPT-3.5-Turbo-1106, and ``GPT-4'' refers to GPT-4-0613. We set the decoding temperature as 0 to maintain the reproducibility of the responses generated by LLMs. All prompts are listed in Appendix~\ref{app:A}, and more dataset information is shown in Appendix~\ref{app:B}.

\subsection{WikiBio GPT-3 Dataset} 
WikiBio GPT-3 dataset is constructed to evaluate the hallucination of LLMs. Researchers randomly select 238 biographical articles from WikiBio dataset~\cite{lebret-etal-2016-neural}, and utilize the text-davinci-003 to generate new passages. The passages are split into 1,908 sentences, and then manually annotated into three categories: \textit{Major Inaccurate}, \textit{Minor Inaccurate}, and \textit{Accurate}.
Following previous studies, \textit{Major Inaccurate} and \textit{Minor Inaccurate} are categorized as \textbf{Nonfactual} (potentially with hallucinations, 1,392 segments), and  \textit{Accurate} is treated as \textbf{Factual} (516 segments).

\textbf{Baselines}: 1)~HalluDetector~\cite{wang-etal-2023-hallucination} utilizes external knowledge sources, a specific classification model and a Naive Bayes classifier to detect hallucination. 2)~Focus~\cite{zhang-etal-2023-enhancing-uncertainty} adopts a multi-stage decision-making process, where both pre-retrieval and task specific classifiers are adopted. 3)~SelfCheckGPT~\footnote{The latest version in https://arxiv.org/pdf/2303.08896.pdf.}, three variants of which are utilized: w/BERTScore, w/Prompt, and w/NLI~\cite{manakul-etal-2023-selfcheckgpt}. SelfCheckGPT w/BERTScore is based on the inherent uncertainty of LLM, while SelfCheckGPT w/Prompt and w/NLI draw upon external knowledge sources. 
The Area Under the Precision-Recall Curve (AUC-PR) and Balanced\_Accuracy are adopted as evaluation metrics. 

\subsection{FELM Dataset}
FELM dataset is designed to evaluate hallucination detection ability. Researchers assemble prompts from diverse scenarios, and use them to instruct GPT-3.5-Turbo-0301 to generate responses. These responses are manually annotated as nonfactual and factual, along with supporting documents. Our experiments are conducted on the WorldKnowledge subset of FELM as it is based on Wikipedia corpus.

\textbf{Baselines}: Following settings in FELM, we adopt four strategies with ChatGPT, GPT-4, and Vicuna-33B as the backbone LLM~\cite{zheng2023judging}: 1) Vanilla prompts. 2) Prompts augmented with Chain-of-Thought (CoT) reasoning~\cite{kojima2022large}. 3) Prompts augmented with hyperlinks to reference documents and 4) Prompts augmented with human-annotated reference documents~\cite{chen2023felm}. Experiments are conducted at the individual response level. Following the previous work~\cite{chen2023felm}, we chose accuracy of nonfactual, factual, and balanced as final metrics to to facilitate comparison with previous works.

\begin{table*}[ht]
\centering
\FloatBarrier
\begin{tabular}{ccccc}
\hline
\multicolumn{2}{c}{\multirow{2}{*}{Method}}                                            & \multicolumn{2}{c}{AUC-PR (\%)}      & \multirow{2}{*}{Balanced\_Acc (\%)} \\ \cline{3-4}
\multicolumn{2}{c}{}           & Nonfactual        & Factual        &                                \\ \hline
\multirow{2}{*}{HalluDetector} & $\mathit{C}_\textit{M}=14, \mathit{C}_\textit{FA}=24$ & 82.42          & 57.01          & 70.54                               \\
                               & $\mathit{C}_\textit{M}=28, \mathit{C}_\textit{FA}=96$ & 86.45          & 61.96          &   \underline{74.82}                           \\ \hline
\multirow{2}{*}{Focus}         & $LLaMA-30B_{\textit{focus}}$                          & 89.79          & 65.69          &       73.64                        \\
                               & $LLaMA-65B_{\textit{focus}}$                          & 89.94          & 64.90          & 74.08                             \\ \hline
\multirow{3}{*}{SelfCheckGPT}  & w/BERTScore                                           & 81.96          & 44.23          & 59.31                              \\
                               & w/NLI                                                 & \textbf{92.50} & 58.47          & 70.55                              \\
                               & w/Prompt                                         & 91.16          & \underline{68.37}          & 72.64                          \\ \hline
CEG                           & top-$k$=6                                                & \underline{92.31}          & \textbf{70.24} & \textbf{77.59}                 \\ \hline
\end{tabular}
\caption{Experimental results of our method powered by GPT-3.5-Turbo-Instruct on WikiBio GPT-3. The SelfCheckGPT with Prompt is also powered by GPT-3.5-Turbo-Instruct because GPT-3.5-Turbo-0613 is deprecated.}
\label{tab:1}
\end{table*}

\subsection{HaluEval Dataset} 

HaluEval dataset is a benchmark for assessing the ability of LLMs to discern hallucinations. Each instance comprises a question, a correct answer, and a hallucinated answer (multiple answers are automatically generated,  and the most confusing one is selected by ChatGPT). 
The QA subset of HaluEval is adopted as it is constructed by Wikipedia corpus, and 2,000 samples of which are randomly sampled.

\textbf{Baselines}: We adopt several models building upon the updated version of ChatGPT as previous studies~\cite{li-etal-2023-halueval}:  1) Vanilla prompts. 2) Prompts augmented with CoT reasoning. 3) Prompts with Pre-RAG, where a strong and fine-tuned retriever, All-mpnet-base-v2\footnote{https://huggingface.co/sentence-transformers/all-mpnet-base-v2.}, is used. Accuracy is chosen as the evaluation metric.

\subsection{WikiRetr Datasets} 
WikiRetr datasets are designed to conduct further analyses on our CEG framework, which is created based on the October 20, 2023 snapshot of Wikipedia. We randomly select 1,000 passages, and apply text-davinci-003 and GPT-4 to rewrite them as new claims, separately. So that each rewritten claim is accompanied by an original passage. The constructed datasets are named  WikiRetr-GPT3 and WikiRetr-GPT4, respectively. Discussion about the reliability of WikiRetr datasets is provided in Appendix~\ref{app:D}.

To analyze the retrieval module, we utilize various retrievers, including: 1) SimCSE BERT, which is employed in our CEG framework; 2) Sentence BERT~\cite{reimers-gurevych-2019-sentence}, a retriever trained with siamese networks; and 3) All-mpnet-base-v2.
Recall@$k$ is the metric to verify if the original document is retrieved in top-$k$. Precision@$k$ is the metric to verify if the claim is supported by a doc in top-$k$ docs with NLI method. 

For NLI methods in the citation generation module, we randomly select 100 instances from each dataset for evaluation. We conduct manual annotation to assess whether the original passages can support the rewritten claims by three annotators. Labeling results show that 90\% and 94\% generated claims are supported by original documents, which is the ground truth of NLI models. Then, we use 1) True-9B\footnote{https://huggingface.co/google/t5\_xxl\_true\_nli\_mixture.}~\cite{honovich-etal-2022-true-evaluating} model and 2) GPT models as NLI methods in analyses. We choose the consistency between manual and model annotations as the evaluation metric.

\section{Experimental Results and Analyses}
\subsection{Performance on Hallucination Detection } 

To verify the effectiveness of our method, we utilize our retrieval augmentation and citation generation modules for hallucination detection on WikiBio GPT-3 and FELM datasets. 

Overall performances in WikiBio GPT-3 dataset are shown in Table~\ref{tab:1}, and we have the following observations: 1) Our CEG framework outperforms all baseline methods in Balanced\_ACC and AUC-PR of Factual segments, achieving the second performance in AUC-PR of nonfactual segments. These results indicate the strength of our method. 2) Previous pre-retrieval augmented generation models, SelfCheckGPT w/NLI and w/prompt, also get good performances in AUC-PR. While suffering from the lengthy text, they cannot achieve comparable performance of our model in all metrics. 3) Our model performs slightly worse than w/NLI in the AUC-PR of nonfactual segments, the reason can be the NLI module of SeftCheckGPT undergoes additional training on detecting nonfactual segments (but achieve poor results in factual).

\begin{table}[t]
\centering
\resizebox{\linewidth}{!}{
\begin{tabular}{ccccc}
\hline
\multicolumn{2}{c}{\multirow{2}{*}{Method}} & \multicolumn{3}{c}{Accuracy (\%)}                       \\ \cline{3-5} 
\multicolumn{2}{c}{}                        & Nonfact & Factual & Balanced                       \\ \hline
\multirow{3}{*}{Vanilla}   & Vicuna-33B     & 72.8    & 34.0    & 53.4                           \\
                           & ChatGPT        & 3.4     & \underline{96.1}    & 49.8                           \\
                           & GPT-4          & 21.8    & 95.3    & 58.5                           \\ \hline
\multirow{3}{*}{CoT}       & Vicuna-33B     & 40.8    & 62.3    & 51.6                           \\
                           & ChatGPT        & 2.7     & \textbf{96.9}    & 49.8                           \\
                           & GPT-4          & 42.9    & 94.0    & \underline{68.4}                           \\ \hline
\multirow{3}{*}{Link}      & Vicuna-33B     & 70.7    & 29.9    & 50.3                           \\
                           & ChatGPT        & 11.6     & 94.3    & 52.9                           \\
                           & GPT-4          & 35.4    & 93.2    & 64.3                           \\ \hline
\multirow{3}{*}{Doc}       & Vicuna-33B     & \underline{81.6}    & 17.9    & 49.8                           \\
                           & ChatGPT        & 34.7    & 73.2    & {54.0}                           \\
                           & GPT-4          & \textbf{88.3}    & 40.8    & 64.6                           \\ \hline
\multirow{3}{*}{CEG}      & Vicuna-33B  & 8.8    & 95.1    & 52.0                           \\
& ChatGPT  & 40.1    & 79.2    & {59.7}                           \\
                           & GPT-4           & 54.4    & 85.5    & \textbf{69.9} \\ \hline
\end{tabular}
}
\caption{Experimental results of our method powered by ChatGPT and GPT-4 on FELM worldknowledge subset. Baselines use the same GPT versions as our CEG, so the performances may vary from their original papers.}
\label{tab:2}
\end{table}

Experimental results in FELM dataset are summarized in Table~\ref{tab:2}. Firstly, our model achieves the best result in balanced accuracy with GPT-4, indicating its effectiveness. Most baseline models are biased in classifying a single type. Then, CEG with ChatGPT beats other ChatGPT baselines, showing the flexibility of our model. Thirdly, CEG outperforms all pre-retrieval baselines, which shows the strength of the proposed post-hoc segment-level retrieval module in hallucination detection. Finally, for Vicuna-33B, all methods exhibit some degree of decline compared to the Vallina method, indicating its limitations in general ability and using retrieved documents. However, our method shows the smallest decline, especially compared to the manually labeled Doc baseline, our method outperforms by 2.2 points, proving the effectiveness of our finer-grained document utilization.


To summarize, CEG outperforms various SOTA baselines in two benchmarks with distinct LLM backbones, indicating that post-hoc retrieval with NLI is powerful for hallucination detection.

\subsection{Results on Hallucination Regeneration} 
\begin{table}[H]
\centering
\scalebox{0.8}{
\begin{tabular}{ccc}
\hline
\multicolumn{2}{c}{Method}                                                 & Accuracy (\%)                                                                                                               \\ \hline
\multicolumn{1}{c}{\multirow{3}{*}{Baselines}} & Vanilla                  & 63.40                                                                                                                  \\
\multicolumn{1}{c}{}                           & w/CoT                    & 68.55                                                                                                                  \\
\multicolumn{1}{c}{}                           & w/Pre-Retrieval          & 61.35                                                                                                                  \\ \hline
\multicolumn{1}{c}{\multirow{2}{*}{CEG}}      & w/ChatGPT                & \underline{69.00}                                                                                                                  \\
\multicolumn{1}{c}{}                           & w/GPT-3.5-Turbo-Instruct & \textbf{69.45} 
\\ \hline
\end{tabular}
}
\caption{Experimental results powered by ChatGPT on the HaluEval QA subset. We employ two GPT models as the NLI method in our citation generation module.}
\label{tab:3}
\end{table}

On the HaluEval dataset, we adopt the full framework of CEG to further evaluate the regeneration module. If the doc is helpful in solving the problem and any of the response segments are classified into nonfactual, our method will generate a new prompt for regeneration as introduced in Section~\ref{sec:3.4}. Besides, a 2018 Wikipedia snapshot is adopted as the corpus~\cite{gao-etal-2023-enabling} in this experiment due to the inconsistency between this dataset and the previous corpus.

Experimental results are presented in Table~\ref{tab:3}. Firstly, our CEG framework with GPT-3.5-Turbo-Instruct achieves the best performance (69.45\% in accuracy), which achieves 8.10\% improvements compared to the pre-hoc retrieval strategy. CEG with ChatGPT also outperforms the pre-retrieval strategy, so these results demonstrate our post-hoc method is robust. 
Secondly, pre-hoc retrieval strategy even performs worse than the baseline with CoT~\cite{li-etal-2023-halueval}, which indicates the retrieved documents are not always helpful. 
Thirdly, consistent with our experiments related to NLI models in Table 5, when using GPT-3.5-Turbo-Instruct as the NLI model, the results are  superior to ChatGPT.
We also conduct case studies to show our regeneration results, and some cases are shown in Appendix~\ref{app:C}.



\subsection{Further Analyses}
\subsubsection{Ablation Study}

\begin{table}[H]
\centering
\resizebox{\linewidth}{!}{
\begin{tabular}{ccccc}
\hline
\multicolumn{2}{c}{\multirow{2}{*}{Variants}} & \multicolumn{3}{c}{Accuracy (\%)} \\ \cline{3-5} 
\multicolumn{2}{c}{}       & Nonfact & Factual & Balanced \\ \hline
\multirow{4}{*}{ChatGPT}   & w/o RA        & 17.7    & 90.1    & 53.9     \\
                           & w/o Threshold  & 40.1    & 78.2    & 59.2     \\
                           & ALL           & 40.1    & 79.2    & 59.7     \\ \hline
\multirow{4}{*}{GPT-4}     & w/o RA        & 30.6    & 93.3    & 61.9     \\
                           & w/o Threshold  & 50.3    & 83.6    & 67.0     \\
                           & ALL           & 54.4    & 85.5    & 69.9     \\ \hline
\end{tabular}
}
\caption{Ablation results of CEG on the Worldknowledge subset of FELM. `RA' means the Retrieval Augmentation module.}
\label{tab:4}
\end{table}

We conduct ablation experiments on the FELM Worldknowledge dataset, where the retrieval augmentation ($k$ = 4) and the document selection threshold are involved (threshold = 0.5). 
As shown in Table~\ref{tab:4}, the retrieval augmentation module plays an important role in providing better results on different backbone LLMs (ChatGPT and GPT-4). Furthermore, the threshold of retrieved documents is necessary, which can filter out irrelevant documents in citation generation. Thus, all designed modules contribute to improvements in the CEG framework. 

\subsubsection{Retrieval Models}
\begin{figure}[H]
    \centering
    \includegraphics[width=0.95\columnwidth]{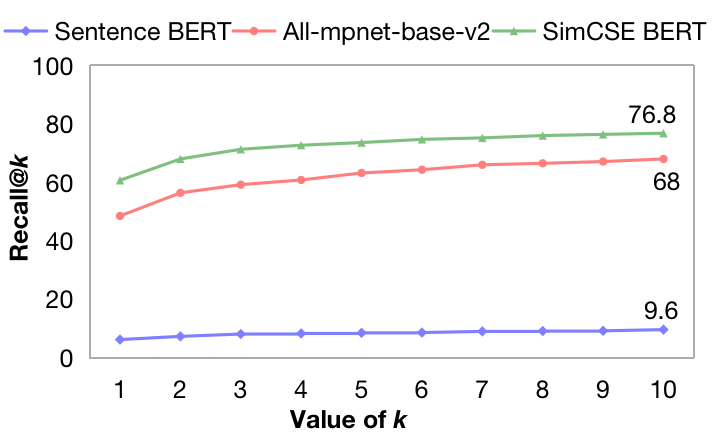}
    \caption{Performances of different retrievers on the WikiRetr-GPT3 dataset.}
    \label{fig:3}
\end{figure}

The choice of retrieval model significantly impacts the performance of our retrieval augmentation module, as illustrated in Figure~\ref{fig:3}. Experimental results of three different retrieval models show that SimCSE BERT has better performances in post-hoc retrieval tasks (76.8\% when using top 10 documents), where 64 million documents are used as candidates. Besides, although a larger value of $k$ can improve recall, it also requires more resources for further computation. For a balance between efficiency and effectiveness, the value of k is set between 4 and 6 in our experiments.

\subsubsection{NLI Models in Citation Generation}
\begin{table}[H]
\centering
\resizebox{\linewidth}{!}{
\begin{tabular}{ccc}
\hline
                 &  WikiRetr-GPT3  & WikiRetr-GPT4 \\\hline
True-9B          & \underline{84}          & 84                    \\
ChatGPT          & 66          & 77                     \\
GPT-3.5-Turbo-Instruct & \textbf{86} & \underline{91}                     \\
GPT-4 Turbo      & 83          & 90                     \\
GPT-4            & 83          & \textbf{96}          \\\hline
\end{tabular}
}
\caption{Agreement rate (\%) of different NLI models with human annotated instances on WikiRetr datasets.}
\label{tab:5}
\end{table}

The performance of different NLI models in the citation generation module is illustrated in Table~\ref{tab:5}, and there are two main observations we can make: 1)~Despite being a state-of-the-art task-specific NLI approach, True-9B performs worse than best LLMs in this scenario. LLMs, such as GPT-3.5-Turbo-Instruct and GPT-4, are capable of playing the NLI model in our citation generation module, as they achieve high agreement rates with human annotators. 2)~LLMs show better performance on data they generate, which is consistent with previous studies~\cite{wang2023openchat,zheng2023judging}.

\begin{table}[ht]
\centering
\resizebox{\linewidth}{!}{
\begin{tabular}{ccccccc}
\hline
Metric     & \multicolumn{4}{c}{Precision (\%)}                           & \multicolumn{2}{c}{Recall (\%)} \\ \hline
NLI model  & \multicolumn{2}{c}{True} & \multicolumn{2}{c}{GPT-4} & \multicolumn{2}{c}{-}         \\ \hline
Top-$k$       & 5          & 10          & 5           & 10          & 5             & 10            \\ \hline
WikiRetr-GPT3 &       71.2     &       74.2      &       \underline{75.2}      &       \textbf{78.2}      &         73.6      & 76.8              \\
WikiRetr-GPT4&      58.1     &      62.6       &       \underline{69.7}      &      \textbf{75.7}       &        57.0       &       60.9        \\ \hline
\end{tabular}
}
\caption{NLI Precision of True-9B and GPT-4 on WikiRetr Datasets.}
\label{tab:6}
\end{table}

Table~\ref{tab:6} shows experimental results of the citation generation module with distinct NLI models when the retriever is Simcse BERT, which indicate: 1) Even on the corpus with over 64 million candidates, our citation generation module exhibits outstanding performance, achieving 78.2 and 75.7 on the precision of the two datasets, respectively. 2) Compared to WikiRetr-GPT3, WikiRetr-GPT4 constitutes a more challenging and higher-quality dataset. The reason is that WikiRetr-GPT4 demonstrates lower recall, suggesting a lower semantic similarity between the original text and the generated claim. While its precision surpasses recall, indicating the generated claims are high quality.

\subsubsection{Hyper-parameter Analysis}

\begin{figure}[H]
    \centering
    \includegraphics[width=0.95\columnwidth]{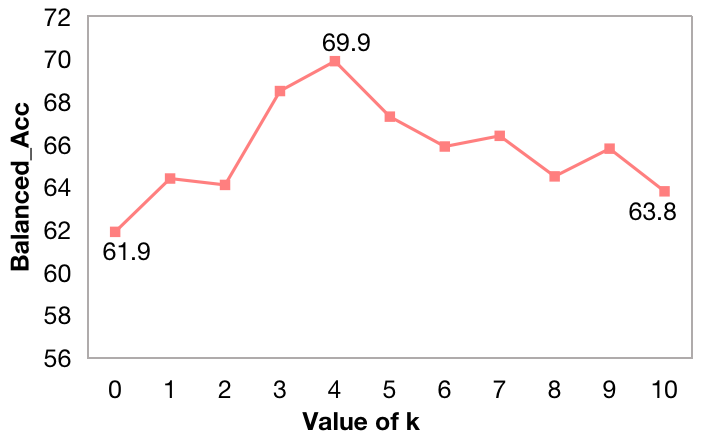}
\caption{The variation of Balanced Accuracy with the change of $k$ on the FELM dataset.}
    \label{fig:5}
\end{figure}


Due to the space limit, we only show the hyper-parameter analysis of $k$ on FELM datasets in Figures~\ref{fig:5}. We have the following observations: 1)~Experimental results on both FELM and WikiBio GPT-3 datasets demonstrate that more documents (larger value of $k$) do not always provide better performance. The reason can be more documents lead to longer input, attention becomes dispersed, and simultaneously, the relevance of the retrieved documents decreases. 2)~Less or no retrieved documents also contribute to worse performance, which indicates the usefulness of the retrieval augmentation module in CEG. 3)~The best performance achieved when $k$ near 5 (Top-$4$ for FELM dataset and Top-$6$ for WikiBio GPT-3 dataset).


\section{Conclusion}
In this study, we propose a novel post-hoc citation-enhanced generation framework to reduce hallucinations in LLMs, which involves retrieval augmentation and natural language inference technologies. 
Different from previous hallucination studies, our framework is post-hoc and flexible, which can be applied to distinct LLMs without additional training or annotations, making it hold significant practical implications.
Experiments on three hallucination-related benchmarks and our datasets demonstrate that our CEG framework achieves state-of-the-art performance in hallucination detection and regeneration. Further analyses show the effectiveness of our proposed modules and adopted models. In the future, we plan to further expand the corpus to support more scenarios.

\section*{Limitations}

Our study has several limitations: 1) Restricted retriever and corpus: In our experiments, we do not employ a fine-tuned specific retriever for post-hoc methods, and our method utilizes only the Wikipedia corpus, limiting the applicability of our framework to general knowledge-based question-answering scenarios but only demonstrate the effectiveness of our model. 2) Our experiments are conducted on existing benchmarks and manual annotations, where we do not propose new QA datasets for the verification of regeneration performance. 3) The adopted NLI method in the citation generation module inherently relies on the LLM's world knowledge. More powerful NLI methods can contribute to better performance. 4) Prompting to regenerate and Using NLI technology to generate citations both incur API cost. API costs incurred for conducting our method and creating the WikiRetr datasets are shown in Table~\ref{tab:19}.

\section*{Acknowledgement}
This work is supported by  the National Natural Science Foundation of China (No. 62276152, 61925601, 62372260). We appreciate all the reviewers for their insightful suggestions.
\bibliography{anthology,custom}
\bibliographystyle{acl_natbib}

\appendix

\section*{Appendix}

\section{Prompts Used in Our Experiments}
\label{app:A}
The prompt used in the evaluation of CEG and w/o threshold variants of CEG on the FELM dataset, as well as in the citation generation experiments on the HaluEval dataset, is presented in Table~\ref{tab:11}. The prompt for w/o retrieval augmentation variants of CEG in the FELM dataset evaluation is presented in Table~\ref{tab:12}.

The prompts for the baseline Vanilla, CoT, and Pre-Retrieval methods used in the HaluEval dataset are shown in Table~\ref{tab:13}, Table~\ref{tab:14}, and Table~\ref{tab:15}. The prompt to regenerate a new response on the HaluEval dataset is presented in Table~\ref{tab:16}.

\noindent\makebox[\linewidth]{\hdashrule{\linewidth}{1pt}{1mm}}
\textbf{Context}: \\
\{ \textit{Retrieved passages} \}                                                          \\
\textbf{Sentence}: \\
\{ \textit{Sentence to be verified} \}                                                        \\
Is the sentence supported by the context above? Answer Yes or No: \\ \noindent\makebox[\linewidth]{\hdashrule{\linewidth}{1pt}{1mm}}

The NLI prompt in experiments on WikiBio dataset is presented above.

\noindent\makebox[\linewidth]{\hdashrule{\linewidth}{1pt}{1mm}}
\textbf{premise}: \{ \textit{Passage retrieved} \} \textbf{hypothesis}: \{ \textit{Hypothesis to be verified} \}\\ 
\noindent\makebox[\linewidth]{\hdashrule{\linewidth}{1pt}{1mm}}

The NLI prompt employed by True-9B in experiments on WikiRetr datasets is presented above. We utilize the same prompt\footnote{You can find the original prompt at https://github.com/princeton-nlp/ALCE/blob/main/eval.py} as in \citealp{gao-etal-2023-enabling}. The agreement rates with human annotaters of True-9B in Table~\ref{tab:5} and the precision of True-9B in Table~\ref{tab:6} are based on this prompt.

\noindent\makebox[\linewidth]{\hdashrule{\linewidth}{1pt}{1mm}}
\textbf{Premise}: \{ \textit{Passage retrieved} \}\\
\textbf{Hypothesis}: \{ \textit{Hypothesis to be verified} \}\\
\textbf{Task}: Determine the logical relationship between premise and hypothesis.\\
\textbf{Output format}: If you believe the relationship is Entailment, output Entailment; for Contradiction, output Contradiction; for Neutral, output Neutral.\\ 
\noindent\makebox[\linewidth]{\hdashrule{\linewidth}{1pt}{1mm}}

The NLI prompt employed by GPT series models in experiments on WikiRetr datasets is presented above. The agreement rates with human annotaters of GPT models in Table~\ref{tab:5} and the precision of GPT models in Table~\ref{tab:6} are based on this.

\section{Detailed Information About Adopted  Datasets}
\label{app:B}
More Statistics of WikiBio GPT-3 dataset are shown in Table~\ref{tab:7}, and statistics of WorldKnowledge subset in FELM are summarized in Table~\ref{tab:8}.
 \begin{table}[H]
 \centering
 \scalebox{0.9}{
 \begin{tabular}{cccc}
 \hline
 \#Passages & \#Sentences & \#Factual & \#Nonfactual \\ \hline
 238        & 1,908        & 516        & 1,392       \\ \hline
 \end{tabular}

 }
 \caption{Statistics of WikiBio GPT-3 dataset.}
 \label{tab:7}
 \end{table}
 \begin{table}[H]
\centering
\resizebox{\linewidth}{!}{
\begin{tabular}{cccc}
\hline
 & \#Response  & Error rate (\%) & agreement rate (\%) \\ \hline
Statistics     & 184       & 46.2            & 81.5            \\ \hline
 & \#Segment & \#Factual    & \#Nonfactual    \\ \hline
Statistics      & 532       & 385             & 147             \\ \hline
\end{tabular}

}
\caption{Statistics of WorldKnowledge subset in FELM. The ``Error rate'' indicates the proportion of responses containing factual errors. The agreement rate is agreement rate between the two annotators during the annotation process.}
\label{tab:8}
\end{table}
In Table~\ref{tab:9}, we provide an example of HaluEval dataset.
\FloatBarrier
\begin{table}[H]
\centering
\scalebox{0.85}{
\begin{tabular}{p{1.1\linewidth}}
\hline
\#Knowledge\#: The nine-mile byway starts south of Morehead, Kentucky and can be accessed by U.S. Highway 60. Morehead is a home rule-class city located along US 60 (the historic Midland Trail) and Interstate 64 in Rowan County, Kentucky, in the United States. \\
\#Question\#: What U.S Highway gives access to Zilpo Road, and is also known as Midland Trail? \\
\hdashline
\#Right Answer\#: U.S. Highway 60 \\
\#Hallucinated Answer\#: U.S. Highway 70 \\
\hline
\end{tabular}

}
\caption{An example of HaluEval dataset.}
\label{tab:9}
\end{table}
The utilized modules in different datasets are summarized in Table~\ref{tab:10}.
\begin{table}[H]
\centering
\resizebox{\linewidth}{!}{
\begin{tabular}{cccc}
\hline
Datasets      &  Retrieval & Citation & Regeneration \\ \hline
WikiBio GPT-3          &           \checkmark     &  \checkmark   &   -         \\
FELM&        \checkmark         &  \checkmark   &      -     \\
HaluEval      &            \checkmark     &   \checkmark  &      \checkmark      \\ 
WikiRetr      &     \checkmark     &  \checkmark   &           - \\ \hline
\end{tabular}
}
\caption{Modules of our framework used in four experimental datasets. "\checkmark" indicates this module is adopted in the experiment, and "-" indicates not used. }
\label{tab:10}
\end{table}

\section{Case Studies on HaluEval Dataset}
\label{app:C}
We conduct case studies on two cases in the evaluation of  the HaluEval dataset, presented in Table~\ref{tab:17} and Table~\ref{tab:18}, respectively.

\section{Discussion about the reliability of WikiRetr datasets}
\label{app:D}
After constructing WikiRetr datasets, we randomly select 100 samples from each of the two datasets, and conduct manual annotation to assess whether the original passages can support the rewritten claims by three annotators. More specifically, each sample is initially annotated by two human annotators. In cases where there are discrepancies between the annotations provided by the two annotators, the final label is determined by consensus among three annotators.

Labeling results show that 90\% and 94\% of generated claims are supported by original documents. This consistency is exceptionally high. For example, in the FELM dataset, the average inter-annotator agreement for labels is 90.7\%, and in the Pinocchio dataset~\cite{hu2023large}, the average label accuracy for sampled data is 92.7\%, and the inter-annotator agreement is 85.6\%.

\begin{table*}[]
\centering
\begin{tabular}{ccc}
\hline
Experiments                          & GPT-3.5 & GPT-4 \\ \hline
Main result on FELM                  & $\sim700$     & $\sim700$   \\
Main result on WikiBio GPT-3         & $\sim2000$    & -     \\
Main result on HaluEval              & $\sim7700$    & -     \\
Variants of CEG on FELM              & $\sim1400$    & $\sim1400$  \\
Top-k ablation on FELM               & $\sim6300$    & $\sim6300$  \\
Top-k ablation on WikiBio GPT-3      & $\sim12000$   & -     \\
Creating WikiRetr datasets           & $\sim1000$    & $\sim1000$  \\
NLI experiments on WikiRetr datasets & -       & $\sim20000$ \\
Annotation on WikiRetr datasets      & $\sim400$     & $\sim400$   \\ \hline
\end{tabular}
\caption{API costs incurred for conducting our method and creating the WikiRetr datasets. We report the number of calls for different GPT models. For GPT-3.5, the total number of calls includes both GPT-3.5-Turbo-1106 and GPT-3.5-Turbo-Instruct. For GPT-4, the total number of calls includes both GPT-4-0613 and GPT-4-1106-preview.}
\label{tab:19}
\end{table*}
\begin{table*}[]
\begin{tabular}{p{1\linewidth}}
\hline
\textbf{Instruction}: I will show you a question, a response segment of this question, and a reference doc. Your task is to assess whether the given response segment contains factual errors or not with the help of the reference doc. If you believe the segment contains factual errors, your answer should be ``Nonfactual''; if there is no factual error in this segment, your answer should be ``Factual''. This means that the answer is ``Nonfactual'' only if there are some factual errors in the response segment. When there is no factual judgment in the response segment or the response segment has no clear meaning, your answer should be ``Factual''. Think it step by step. Give your reasoning first and then output the Answer. \\
\textbf{Question}:\\
\{ \textit{Question} \}\\
\textbf{Response segment}:\\
\{ \textit{Response segment to be verified} \}\\
\textbf{Reference doc}:\\
\{ \textit{Top-k docs concatenated with newline characters and numbers.} \}\\

Output:\\\hline
\end{tabular}
\caption{Prompt used for the evaluation of CEG and w/o threshold variants of CEG on the FELM dataset, as well as in the NLI experiments on the HaluEval dataset.}
\label{tab:11}
\end{table*}
\begin{table*}[]
\begin{tabular}{p{1\linewidth}}
\hline
\textbf{Instruction}: I will show you a question, a response segment of this question. Your task is to assess whether the given response segment contains factual errors or not. If you believe the segment contains factual errors, your answer should be ``Nonfactual''; if there is no factual error in this segment, your answer should be ``Factual''. This means that the answer is ``Nonfactual'' only if there are some factual errors in the response segment.When there is no factual judgment in the response segment or the response segment has no clear meaning, your answer should be ``Factual''. Think it step by step. Give your reasoning first and then output the Answer.\\
\textbf{Question}:\\
\{ \textit{Question} \}\\
\textbf{Response segment}:\\
\{ \textit{Response segment to be verified} \}\\
\textbf{Output}:\\\hline
\end{tabular}
\caption{Prompt used for the evaluation of w/o RA variant of our method on the FELM dataset.}
\label{tab:12}
\end{table*}

\begin{table*}[]

\begin{tabular}{p{1\linewidth}}
\hline
\textbf{Instruction}: I want you act as an answer judge. Given a question, two answers, your objective is to select the best and correct answer without hallucination and nonfactual information. You should try your best to select the best and correct answer. If the two answers are the same, you can randomly choose one. If both answers are incorrect, choose the better one. You MUST select an answer from the provided two answers. Your response should be ``Answer 1'' or ``Answer 2''.\\
\textbf{\#Question\#}: \{ \textit{Question} \} \\
\textbf{\#Answer 1\#}: \{ \textit{Right\_answer} \} \\
\textbf{\#Answer 2\#}: \{ \textit{Hallucinated\_answer} \}\\\hline
\end{tabular}
\caption{Prompt for the baseline Vanilla used in the HaluEval dataset.}
\label{tab:13}
\end{table*}

\begin{table*}[]

\begin{tabular}{p{1\linewidth}}
\hline
\textbf{Instruction}: I want you act as an answer judge. Given a question, two answers, your objective is to select the best and correct answer without hallucination and nonfactual information. You should try your best to select the best and correct answer. If the two answers are the same, you can randomly choose one. If both answers are incorrect, choose the better one. You MUST select an answer from the provided two answers. Think it step by step. Give your reasoning first and then output your choice. Output in the following format, ``\#Reasoning\#:Your Reasoning\textbackslash{}n\#Choice\#:Your Choice''. Your choice should be ``Answer 1'' or ``Answer 2''.\\
\textbf{\#Question\#}: \{ \textit{Question} \} \\
\textbf{\#Answer 1\#}: \{ \textit{Right\_answer} \} \\
\textbf{\#Answer 2\#}: \{ \textit{Hallucinated\_answer} \}\\\hline
\end{tabular}
\caption{Prompt for the baseline CoT used in the HaluEval dataset.}
\label{tab:14}
\end{table*}

\begin{table*}[]

\begin{tabular}{p{1\linewidth}}
\hline
\textbf{Instruction}: I want you act as an answer judge. Given a question, two answers, and related knowledge, your objective is to select the best and correct answer without hallucination and non-factual information. You should try your best to select the best and correct answer. If the two answers are the same, you can randomly choose one. If both answers are incorrect, choose the better one. You MUST select an answer from the provided two answers. Think it step by step. Give your reasoning first and then output your choice. Output in the following format, ``\#Reasoning\#:Your Reasoning\textbackslash{}n\#Choice\#:``X''''. ``X'' should be ``Answer 1'' or ``Answer 2''.\\
\textbf{\#Question\#}: \{ \textit{Question} \} \\
\textbf{\#Answer 1\#}: \{ \textit{Right\_answer} \} \\
\textbf{\#Answer 2\#}: \{ \textit{Hallucinated\_answer} \}\\\hline
\end{tabular}
\caption{Prompt for the baseline Pre-Retrieval used in the HaluEval dataset.}
\label{tab:15}
\end{table*}

\begin{table*}[]

\begin{tabular}{p{1\linewidth}}
\hline
\textbf{User} (Round 1):\\
Instruction: I want you act as an answer judge. Given a question, two answers, your objective is to select the best and correct answer without hallucination and nonfactual information. You should try your best to select the best and correct answer. If the two answers are the same, you can randomly choose one. If both answers are incorrect, choose the better one. You MUST select an answer from the provided two answers. Think it step by step. Give your reasoning first and then output your choice. Output in the following format, ``\#Reasoning\#:Your Reasoning\textbackslash{}n\#Choice\#:Your Choice''. Your choice should be ``Answer 1'' or ``Answer 2''.\\
\#Question\#: \{ \textit{Question} \} \\
\#Answer 1\#: \{ \textit{Right\_answer} \} \\
\#Answer 2\#: \{ \textit{Hallucinated\_answer} \} \\
\hdashline
\textbf{Chatbot} (Round 1):\\ 
\textit{Reasoning and Answer}\\
\hdashline
\textbf{User} (Round 2): \\
Documents:\\ 
\{ \textit{Docs retrieved for the nonfactual segments} \}\\
In your previous response, there are factual inaccuracies in the following passages:\\
\{ \textit{Nonfactual segments} \}\\
Please re-answer the previous question with the help of documents. If the documents are unrelated to the issue, please ignore them. Output in the following format, ``\#Reasoning\#:Your Reasoning\textbackslash{}n\#Choice\#:Your Choice''. Your choice should be ``Answer 1'' or ``Answer 2''.\\ \hline
\end{tabular}
\caption{Prompt used for regeneration of our method on the HaluEval dataset.}
\label{tab:16}
\end{table*}
\begin{table*}
\centering
\begin{tabular}{p{1\linewidth}}
\hline
\rowcolor[HTML]{FFFFC7} 
\textbf{\#Question\#}: What film was written by the newspaperman born in 1880? \\
\rowcolor[HTML]{FFFFC7} 
\textbf{\#Right\_Answer\#}: The Lemon Drop Kid\\
\rowcolor[HTML]{FFFFC7} 
\textbf{\#Hallucinated\_Answer\#}: The Great American Novel written by Damon Runyon.\\ \hline
\rowcolor[HTML]{FFCCC9}\textbf{\#Reasoning 1\#}: Damon Runyon was a newspaperman born in 1880 and is known for writing The Great American Novel.\\
\rowcolor[HTML]{FFCCC9}\textbf{\#Choice 1\#}: Answer 2: The Great American Novel written by Damon Runyon.\\
\rowcolor[HTML]{FFCCC9} {\textbf{\#Segments\#}: 1.Damon Runyon was a newspaperman born in 1880 and is known for writing The Great American Novel.}                                                                                                                                                                                                                                                                                                                                                                                                                                                                                                                                                      \\
\rowcolor[HTML]{FFCCC9} 
{\color[HTML]{000000} \textbf{\#NLI Results\#}: {[}0{]}}                                                                                                                                                                                                                                                                                                                                                                                                                                                                                                                                                                                                                                                                                                            \\
\rowcolor[HTML]{FFCCC9} 
{\color[HTML]{000000} \textbf{\#Doc Retrieved for Segment 1\#}: Damon Runyon Alfred Damon Runyon (October 4, 1880 – December 10, 1946) was an American newspaperman and short-story writer. He was best known for his short stories celebrating the world of Broadway in New York City that grew out of the Prohibition era. To New Yorkers of his generation, a ``Damon Runyon character'' evoked a distinctive social type from the Brooklyn or Midtown demi-monde. The adjective ``Runyonesque'' refers to this type of character as well as to the type of situations and dialog that Runyon depicted. He spun humorous and sentimental tales of gamblers, hustlers, actors, and gangsters, few of} \\ \hline
\rowcolor[HTML]{CDE6C7}\textbf{\#Reasoning 2\#}: Damon Runyon was known for his short stories celebrating the world of Broadway in New York City, not for writing The Great American Novel.                                                                                                                                                                                                                                                                                                                                                                                                                                                                                                                                                 \\
\rowcolor[HTML]{CDE6C7} \textbf{\#\#}: Answer 1: The Lemon Drop Kid                                                                                                                                                                                                                                                                                                                                                                                                                                                                                                                                                                                                                                                                                               \\ \hline
\end{tabular}
\caption{HaluEval Example 1. The \colorbox[HTML]{FFCCC9}{red} part is the content generated for the first time and annotated with citations. The \colorbox[HTML]{CDE6C7}{green} part is the content of the regeneration.}
\label{tab:17}
\end{table*}
\begin{table*}
\centering
\begin{tabular}{p{1\linewidth}}
\hline
\rowcolor[HTML]{FFFFC7} 
\textbf{\#Question\#}: Which low-cost airline is headquartered in Sydney? \\
\rowcolor[HTML]{FFFFC7} 
\textbf{\#Right\_Answer\#}: Jetstar Airways\\
\rowcolor[HTML]{FFFFC7} 
\textbf{\#Hallucinated\_Answer\#}: Tiger Airways is headquartered in Sydney.\\ \hline
\rowcolor[HTML]{FFCCC9}\textbf{\#Reasoning 1\#}: Jetstar Airways is a well-known low-cost airline and is headquartered in Melbourne, not Sydney. Tiger Airways used to be headquartered in Singapore, not Sydney.\\
\rowcolor[HTML]{FFCCC9}\textbf{\#Choice 1\#}: Answer 2: Tiger Airways is not headquartered in Sydney.\\
\rowcolor[HTML]{FFCCC9} {\textbf{\#Segments\#}: 1. Jetstar Airways is a well-known low-cost airline and is headquartered in Melbourne, not Sydney. 2. Tiger Airways used to be headquartered in Singapore, not Sydney}                                                                                                                                                                                                                                                                                                                                                                                                                                                                                                                                                      \\
\rowcolor[HTML]{FFCCC9} 
{\color[HTML]{000000} \textbf{\#NLI Results\#}: {[}0,0{]}}                                                                                                                                                                                                                                                                                                                                                                                                                                                                                                                                                                                                                                                                                                            \\
\rowcolor[HTML]{FFCCC9} 
{\color[HTML]{000000} \textbf{\#Doc Retrieved for Segment 1\#}: Jetstar Airways Jetstar Airways Pty Ltd, trading as Jetstar, is an Australian low-cost airline (self-described as \textbackslash{}"value based\textbackslash{}") headquartered in Melbourne. It is a wholly owned subsidiary of Qantas, created in response to the threat posed by low-cost airline Virgin Blue. Jetstar is part of Qantas' two brand strategy of having Qantas Airways for the premium full-service market and Jetstar for the low-cost market. Jetstar carries 8.5\% of all passengers travelling in and out of Australia. The airline operates an extensive domestic network as well as regional and international services from its main base at Melbourne Airport, using a mixed} \\
\rowcolor[HTML]{FFCCC9} 
{\color[HTML]{000000} \textbf{\#Doc Retrieved for Segment 2\#}: Australia Asia Airlines Australia Asia Airlines\textbackslash{}``Àoyà Hángkōng Gōngsī\textbackslash{}'') was a wholly owned subsidiary of Qantas set up to operate services between Australia and Taiwan (Republic of China). The subsidiary was established due to the People's Republic of China objection to national carriers of countries with which it had diplomatic relations flying to a territory that it regarded as a breakaway province. The airline operated two Boeing 747SPs and a Boeing 767 aircraft seconded from the Qantas fleet, repainted in a modified livery, which did not display the Flag of Australia, or the kangaroo logo, which was replaced by}                \\ \hline
\rowcolor[HTML]{CDE6C7}\textbf{\#Reasoning 2\#}: According to the provided document, Jetstar Airways is headquartered in Melbourne, not Sydney. There is no mention of Tiger Airways being headquartered in Sydney.                                                                                                                                                                                                                                                                                                                                                                                                                                                                                                                                                   \\
\rowcolor[HTML]{CDE6C7} \textbf{\#Choice 2\#}: Answer 1: Jetstar Airways                                                                                                                                                                                                                                                                                                                                                                                                                                                                                                                                                                                                                                                                                               \\ \hline
\end{tabular}
\caption{HaluEval Example 2. The \colorbox[HTML]{FFCCC9}{red} part is the content generated for the first time and annotated with citations. The \colorbox[HTML]{CDE6C7}{green} part is the content of the regeneration.}
\label{tab:18}
\end{table*}

\end{document}